\documentclass[letterpaper]{article} 
\usepackage{aaai23}  
\usepackage{times}  
\usepackage{helvet}  
\usepackage{courier}  
\usepackage[hyphens]{url}  
\usepackage{graphicx} 
\usepackage{natbib}  
\usepackage{caption} 
\usepackage{algorithm}
\usepackage{algorithmic}
\usepackage{amsmath}
\usepackage{makecell}
\usepackage{array}
\usepackage{multirow}
\usepackage{booktabs}
\usepackage{color,soul}
\usepackage[switch]{lineno}
\usepackage[dvipsnames]{xcolor}
\newcommand{\etal}{\textit{et al}.}
\newcommand{\ie}{\textit{i}.\textit{e}.}
\newcommand{\eg}{\textit{e}.\textit{g}.}

\newcommand{\texadd}[1]{\textcolor{black}{#1}}

\newcommand{\zjh}[1]{{#1}}
\newcommand{\whzjh}[1]{{\color{black}#1}}

\usepackage{newfloat}
\usepackage{listings}
\DeclareCaptionStyle{ruled}{labelfont=normalfont,labelsep=colon,strut=off} 
\floatstyle{ruled}
\newfloat{listing}{tb}{lst}{}
\floatname{listing}{Listing}

\title{Hierarchical \whzjh{Consistent} Contrastive Learning for Skeleton-Based Action Recognition with \whzjh{Growing} Augmentations}

\author{
    Jiahang Zhang,
    Lilang Lin,
    Jiaying Liu\thanks{Corresponding author. This work is supported by the National Natural Science Foundation of China under contract No.62172020.}
}
\affiliations{
    {Wangxuan Institute of Computer Technology, Peking University, Beijing, China}\\

    \{zjh2020, linlilang, liujiaying\}@pku.edu.cn
}

\usepackage{bibentry}

\begin{document}

\maketitle
\begin{abstract}
Contrastive learning has been \whzjh{proven} beneficial for self-supervised skeleton-based action recognition.
Most contrastive learning methods \whzjh{utilize} carefully designed augmentations to generate different movement patterns of \whzjh{skeletons} for the same \whzjh{semantics}.
\whzjh{
However, it is still a pending issue to apply strong augmentations, which distort the \whzjh{images/skeletons'} structures and cause semantic loss, due to their resulting unstable training.
}
\whzjh{In this paper, we investigate the potential of adopting strong augmentations and} propose a general hierarchical consistent contrastive learning framework (HiCLR) for skeleton-based action recognition.
Specifically, we first design a \whzjh{gradual growing augmentation policy} to generate multiple ordered positive pairs, \whzjh{which guide to achieve the consistency of the learned representation from different views.}
Then, an asymmetric loss is proposed \whzjh{to enforce  the hierarchical consistency via} \whzjh{a directional clustering operation in the feature space, pulling the representations from strongly augmented views closer to those from weakly augmented views for better generalizability.} 
Meanwhile, we propose and \whzjh{evaluate} three \whzjh{kinds} of strong \whzjh{augmentations} for 3D \whzjh{skeletons} \whzjh{to} demonstrate the effectiveness of our method.
Extensive experiments show that HiCLR \whzjh{outperforms} the state-of-the-art methods notably on three large-scale datasets, \textit{i.e.,} NTU60, NTU120, and PKUMMD. Our project is publicly available at: {https://jhang2020.github.io/Projects/HiCLR/HiCLR.html}.
\end{abstract}
\section{Introduction}

Human action recognition \whzjh{is important} for bridging artificial systems and humans in the real world.
It has been widely used \whzjh{in} video understanding, human-robot interaction, entertainment, \textit{etc.}~\cite{tang2020uncertainty,rodomagoulakis2016multimodal,shotton2011real}. 
Owing to the \whzjh{advantages} such as lightweight, robustness, and privacy protection, action recognition based on 3D skeleton data has attracted \whzjh{a lot of} attention recently. 
There are many works targeted at skeleton-based action recognition~\cite{shi2019two,cheng2020skeleton,liu2020disentangling,chen2021channel}, but most of them are designed in a fully-supervised manner and require \whzjh{a large amount of} labeled data.
Considering that \whzjh{annotation} of large-scale datasets is expensive and time-consuming, recently more and more researchers pay attention to the study of representation learning from unlabeled data~\cite{lin2020ms2l,li20213d,Kim2022GloballocalMT}. 

\begin{figure}[t]
    \centering
    \includegraphics[width=0.97\linewidth]{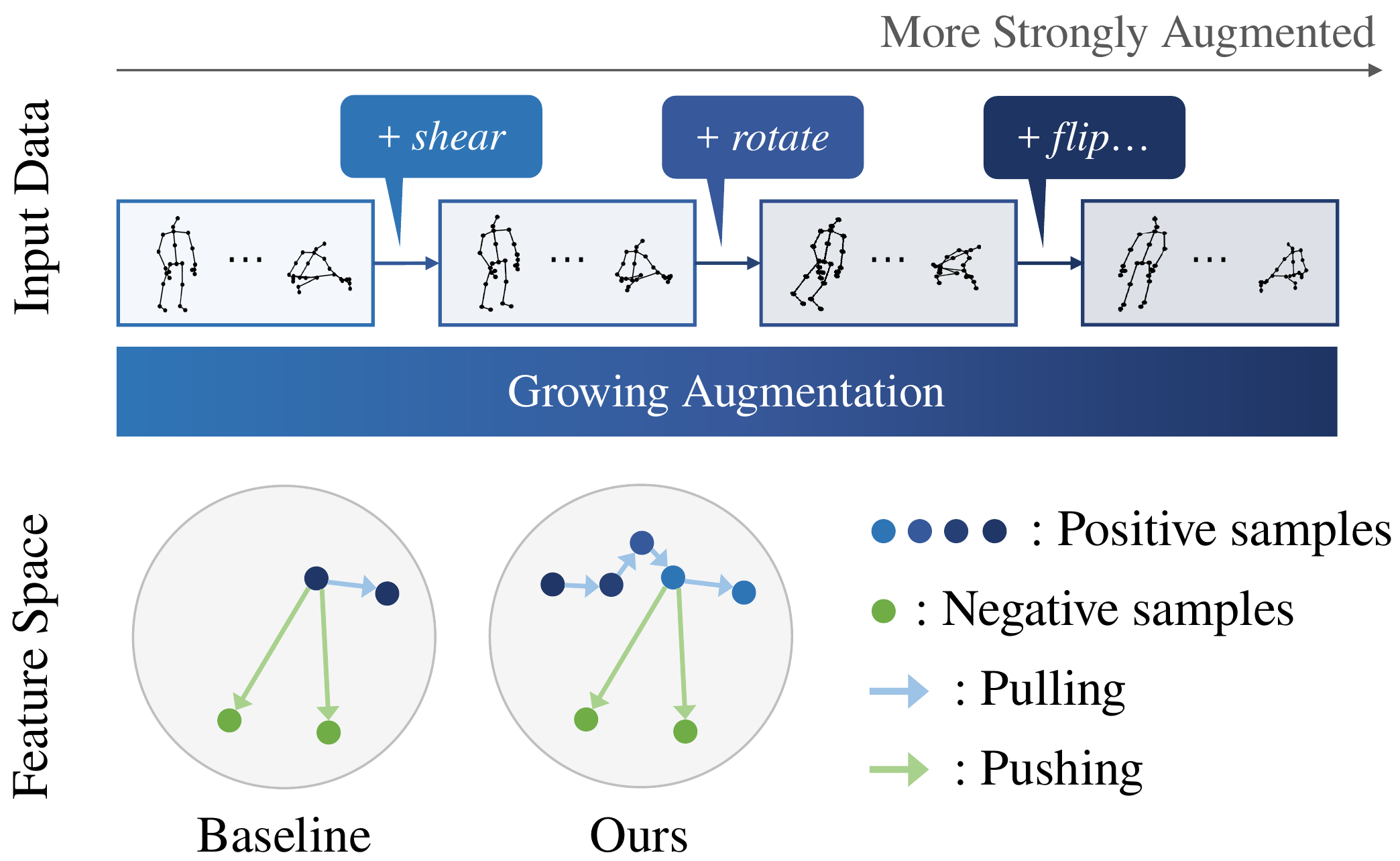}
    \caption{
    The proposed hierarchical consistent contrastive learning compared with the traditional contrastive learning pipelines. 
    Instead of applying all augmentations directly, we utilize \whzjh{a} growing augmentation to generate multiple ordered positive pairs that are augmented progressively.
    Then the model performs a directional feature clustering operation to constrain the consistency of adjacent positive samples.
    }
    \vspace{-5mm}
    \label{fig:teasor}
\end{figure}

Among the various self-supervised learning methods, contrastive learning is an effective \whzjh{one} and has been shown successful for skeleton-based action recognition~\cite{rao2021augmented,thoker2021skeleton,li20213d}.
For contrastive learning, augmentations have been \whzjh{proven} to be very crucial, \whzjh{introducing} various movement patterns for the same \whzjh{semantics} and directly affecting the quality of feature representations learned by the model~\cite{tian2020makes,guo2022aimclr}.
However, \whzjh{it is still not fully investigated on} what augmentation to use and how to use it for skeleton data.

Compared to the RGB representation of human action, 3D skeleton data is a \whzjh{more} high-level modality representation, which \whzjh{intensifies} the sensitivity of contrastive learning to the augmentations. 
\whzjh{This sensitivity leads to a cautious selection of augmentations, which becomes the bottleneck in designing more advanced contrastive learning methods.} 
On the one hand, as shown on the right of Table~\ref{fig:h_loss}, some augmentations like \textit{Random Mask} cause the performance \whzjh{drop} of the baseline algorithm. 
Following~\cite{bai2022directional}, these augmentations are called \textbf{strong augmentations} (also namely \textbf{heavy augmentations}), which distort the images/skeletons' structures and cause semantic loss, leading to unstable training.
Some works have revealed the potential of \whzjh{using} strong data augmentations~\cite{cubuk2020randaugment,wang2021contrastive}. 
However, it \whzjh{is still difficult to measure and constrain the consistency, which is the base of contrastive learning,} directly \whzjh{and accurately} from the strongly augmented views, as these augmentations can cause serious semantic information loss.
On the other hand, \whzjh{most} previous works based on contrastive learning simply treat all augmentations fairly, ignoring the differences in the importance of applied augmentations.
Recent works~\cite{tian2020makes,zhang2022rethinking} have shown that each augmentation has a different impact on \whzjh{the} downstream tasks, and hence \whzjh{learning from the invariance after augmentation without distinction} can inevitably cause non-optimal representations for the downstream task. 

To address the aforementioned issues, we are inspired to explore a general contrastive framework that applies {growing} augmentations. 
In this paper, we propose a novel hierarchical consistent contrastive learning framework (HiCLR) that \whzjh{learns from the invariance of the hierarchical growing augmentation} and \whzjh{treats different augmentations in a distinguished manner}.
Compared to previous works~\cite{li20213d,guo2022aimclr}, we focus on how to better utilize and benefit from \whzjh{multiple augmentations including} strong augmentations.
Specifically, a \whzjh{growing} hierarchical augmentation policy is proposed to construct multiple \whzjh{correlated} positive pairs.
Each of the pairs is generated via more augmentations than the previous one and \whzjh{expands} the feature distribution.
Then, to better utilize the novel patterns brought by the strong augmentations, we propose \whzjh{an} asymmetric hierarchical learning strategy.
Instead of directly learning all augmentation invariances at once, \whzjh{our objective suggests a hierarchical consistent learning manner for different augmentations} as shown in Figure~\ref{fig:teasor}.
Meanwhile, this asymmetric design encourages the strongly augmented view to be similar to the weakly augmented view, \whzjh{which helps the model better generalize.}
\whzjh{Based on our new framework}, we \whzjh{further analyze and evaluate} different choices for strong augmentations. Extensive experiments on both \whzjh{Graph Convolutional Networks (GCNs)} and \whzjh{transformers} are conducted to verify the effectiveness of our method.

Our contributions can be summarized as follows:
\begin{itemize}
\item We propose a hierarchical \whzjh{consistent} contrastive learning framework, HiCLR, which \whzjh{successfully} introduces strong augmentations to the traditional contrastive learning pipelines for skeletons. 
The hierarchical design \whzjh{integrates different augmentations} and \whzjh{alleviates} the difficulty in learning consistency from strongly augmented views, \whzjh{which are accompanied by} serious semantic information loss.

\item We introduce the growing augmentation along with asymmetric hierarchical learning that \whzjh{constrains} the representation consistency of the constructed positive pairs. 
By virtue of these, the model further improves the representation \whzjh{capacity} by leveraging the rich information brought by the strong augmentations.

\item With the proposed framework, we \whzjh{further} design and analyze three strong augmentation strategies: Random Mask, Drop/Add Edges, and SkeleAdaIN. 
Despite the adverse effects observed when applying them directly, they become significantly effective with our HiCLR and exceed state-of-the-art performance.
\end{itemize}

\section{Related Works}
\subsection{Skeleton-based Action Recognition}
Skeleton-based action recognition aims \whzjh{to} classify the action categories using 3D coordinates data of \whzjh{the} human body. 
The current methods can be divided into \whzjh{recurrent neural network (RNN)}-based, \whzjh{convolutional neural network (CNN)}-based, GCN-based\whzjh{,} and \whzjh{transformer}-based styles.
\whzjh{The work in}~\cite{du2015hierarchical} directly uses RNN to tackle the skeleton \whzjh{sequence} data.
Some works~\cite{ke2017new,liu2017enhanced} transform each skeleton sequence into \whzjh{image-like} representations and apply the CNN model to extract the spatial-temporal information. 
Recently, inspired by the natural topology structure of \whzjh{the} human body, GCN-based methods have attracted more attention for skeleton-based action recognition.
Spatial-temporal GCN \whzjh{(ST-GCN)}~\cite{yan2018spatial} firstly \whzjh{explores} the potential of modeling \whzjh{the} spatial-temporal relationship of human skeleton sequences. 
\whzjh{Many} works~\cite{shi2019two,cheng2020skeleton,chen2021channel} based on it have achieved \whzjh{success} by virtue of the GCNs' strong representation \whzjh{capacity}.
Meanwhile, \whzjh{transformer}-based models~\cite{shi2020decoupled,plizzari2021skeleton} also show remarkable results \whzjh{by utilizing the long-range temporal dependencies}, owing to the attention mechanism. 
We adopt the ST-GCN and DSTA-Net~\cite{shi2020decoupled} as our \whzjh{backbones} to evaluate our \whzjh{framework}.

\subsection{Contrastive Representation Learning}
Contrastive learning~\cite{he2020momentum,chen2020simple,chen2020improved} is a popular and effective method for self-supervised learning.
In many works~\cite{tian2020makes,zhang2022rethinking}, the design of augmentations has been found essential for the success of contrastive learning.
\texadd{Xiao \etal~propose Leave-one-out Contrastive Learning~\cite{xiao2020should}, which projects the input image into multiple embedding spaces corresponding to invariance learning of different augmentation combinations. 
The work~\cite{zhang2022hierarchically} proposes to decouple spatial-temporal contrastive learning by applying temporal and spatial augmentations separately.}

Recent works have shown more and more \whzjh{interests} in \whzjh{strong} augmentations. Contrastive learning with
stronger augmentations \whzjh{(CLSA)}~\cite{wang2021contrastive} shows the improvements of strong augmentations in contrastive learning. 
The work \whzjh{in}~\cite{zhang2022rethinking} proposes to apply more augmentations in different depths of the encoder to learn the augmentation invariances \whzjh{non-homogeneously}. 
Bai \etal~\cite{bai2022directional} explore a directional self-supervised objective for heavy image augmentations. 
These works provide an important basis for our research.

For contrastive learning in skeleton-based action recognition, AS-CAL~\cite{rao2021augmented} directly applies the current contrastive learning framework~\cite{he2020momentum} for skeleton. 
Li {\etal}~\cite{li20213d} explores the cross-stream knowledge for contrastive learning. 
Recently, abundant information mining for self-supervised action representation (AimCLR)~\cite{guo2022aimclr} migrates CLSA to skeleton data and uses more augmentations. 
However, these works still lack effective design for the use of strong augmentations, and leave the potential of strong augmentations underutilized. 
To this end, we propose a hierarchical \whzjh{consistent} contrastive learning framework \whzjh{that can effectively borrow the knowledge of strong augmentations}.
\zjh{\section{Proposed Method: HiCLR}}

\subsection{Contrastive Learning for Skeleton}
\label{sec:skeletonclr}
We first give {a unified formulation of the contrastive learning}~\cite{bai2022directional} for skeleton following the recent works:
\begin{itemize}
\item \textbf{Data augmentation module} containing the augmentation strategy set $\mathcal{T}$ to generate the different views of the original data which are regarded as the positive pairs. 
\whzjh{\item \textbf{Query/key encoder $f(\cdot)$} for mapping the input to the latent feature space.
\item \textbf{Embedding projector $h(\cdot)$} for mapping the latent feature into an embedding space where the self-supervised loss is applied.}
\item \textbf{Self-supervised loss} that performs the feature clustering operation in the embedding space.
\end{itemize}

SkeletonCLR~\cite{li20213d} follows the recent contrastive learning framework, MoCov2~\cite{chen2020improved}, and is used as the baseline algorithm of our method. Specifically, given a skeleton sequence $s$, the positive pair $(x, x^{\prime})$ is constructed via $\mathcal{T}$. Subsequently, we can obtain the corresponding feature representations $(z, z^{\prime})$ via the query/key encoder \whzjh{$f(\cdot)$} and embedding projector \whzjh{$h(\cdot)$,} respectively. 
A memory queue \textbf{M} is maintained storing lots of negative samples for contrastive learning. The whole network is optimized by InfoNCE loss~\cite{oord2018representation}:
\begin{equation}\label{eq:infonce}
	\mathcal{L}_{Info} = -\log\frac{\exp(z \cdot z^{\prime} / \tau)}{\exp(z \cdot z^{\prime} / \tau) + \sum_{i=1}^{M}{\exp(z \cdot m_{i} / \tau)}},
\end{equation}
\whzjh{where $m_{i}$ is the feature in \textbf{M} corresponding to the $i$-th negative sample, $M$ is the number of negative features and $\tau$ is the temperature hyper-parameter.}
After each training step, all samples in \whzjh{a} batch will be updated to \textbf{M} as negative samples in a first-in, first-out policy.
The key encoder is a momentum-updated version of the query encoder that is updated via gradients.
Concretely, denoting the parameters of query encoder and key encoder as $\theta_{q}$ and $\theta_{k}$ respectively, the key encoder is updated as:
$\theta_{k} \leftarrow m\theta_{k} + (1-m)\theta_{q},$
where $m \in [0,1)$ is a momentum coefficient. 

\begin{figure*}[t]
    \centering
    \includegraphics[width=0.9\linewidth]{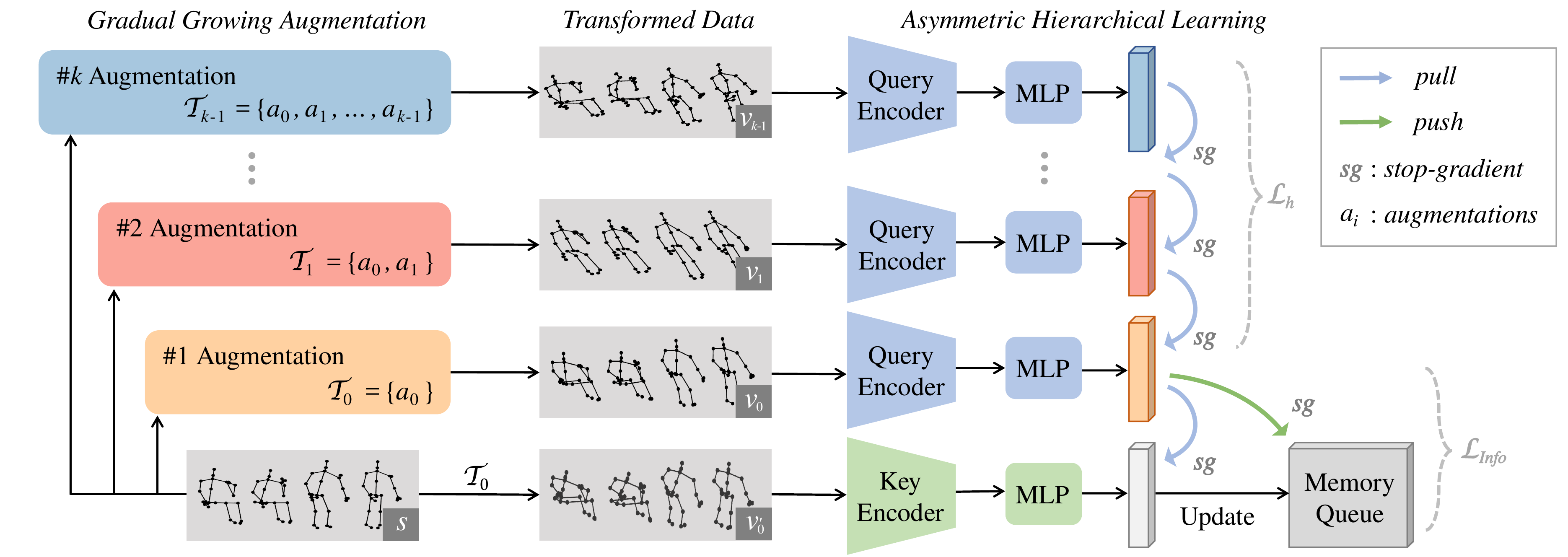}
    \caption{
    The overview architecture of the proposed HiCLR.
    There are \whzjh{$k$} branches sharing the same query encoder weights corresponding to the hierarchical learning of different augmentations. 
    The augmented view $v_{i}$ is fed into the query encoder $f_{\theta_{q}}$ and the \whzjh{embedding projector} $h_{\theta_{q}}$ to obtain $z_{i}$. 
    Similarly, $z^{\prime}_{0}$ is obtained by the key encoder $f_{\theta_{k}}$ and the \whzjh{embedding projector} $h_{\theta_{k}}$. 
    Meanwhile, a hierarchical self-supervised loss is proposed \whzjh{to align the feature distributions of} adjacent branches, which is optimized jointly with the InfoNCE loss.
    }
    \vspace{-5mm}
    \label{fig:architecture}
\end{figure*}

\subsection{Hierarchical \whzjh{Consistent} Contrastive Learning}

\whzjh{Traditional contrastive learning works directly apply the augmentation sets at once to generate positive pairs.
When strong augmentations are applied, these positive samples heavily suffer from semantic information loss, sharing less correlation.
However, it is quite difficult to learn useful information from the consistency constraint of these degraded pairs.
To address this problem, we propose a hierarchical consistent contrastive learning framework.
We generate a series of highly correlated positive pairs progressively via gradually growing augmentations.
Therefore, these pairs provide hierarchical guidance of the feature similarity and benefit the model in learning the knowledge from strong augmentations with consistency of different views.}

We first give an overview of our method. 
As shown in Figure~\ref{fig:architecture}, HiCLR has multiple branches \whzjh{to extract features} and mainly comprises two components:
(1) A gradual \whzjh{growing} augmentation policy which constructs multiple positive pairs corresponding to the different augmentations. 
(2) Asymmetric hierarchical learning \whzjh{constraint} of the representation consistency from strongly augmented views.
Next\whzjh{,} we will introduce each component \whzjh{in detail}.
\vspace{1mm}

\noindent \textbf{\whzjh{1) Gradual growing} augmentation.} 
\whzjh{To facilitate the learning process to achieve better augmentation invariance, 
a gradual growing augmentation policy is introduced.}
\whzjh{The augmentation policy consists of} multiple augmentation sets, each of which is an extended version of the existing one.
By virtue of this, multiple ordered positive pairs with increasing distortion are generated.

Here we give a formal description of our growing augmentation pipeline.
The proposed growing hierarchical augmentation policy constructs the following augmentation sets: $\mathcal{T}_{0}, \mathcal{T}_{1}, ...\ ,\mathcal{T}_{k-1}$, where \whzjh{$k$} is the total number of different augmentation sets. 
$\mathcal{T}_{0}$ contains the basic augmentation strategy\whzjh{,} and each set adopts more augmentation instances than the previous one.
These sets can be formulated as:
\begin{equation}\label{eq:PAM}
\begin{aligned}
	\mathcal{T}_{0} &= \left\{ a_{0,0} \right\}, \\
	\mathcal{T}_{1} &= \left\{ a_{0,1},\  a_{1,1} \right\}, \\
	 &\ ... \\
	\mathcal{T}_{k-1} &= \left\{ a_{0,k-1},\  a_{1,k-1},\  ...\ , \ a_{k-1,k-1} \right\},
\end{aligned}
\end{equation}
where $a_{i,j}$ represents the instances sampled from the $i$-th augmentation strategy belonging to the $j$-th augmentation set. 
Note that we re-sample the instances of each augmentation strategy in each augmentation set, which means that
$a_{i,j}\neq a_{i,j^{\prime}}$, $j\neq j^{\prime}$. 
The re-sampling strategy further expands the feature distribution, enabling the model to learn a more \whzjh{distinguishable} feature space for the downstream task.

Resorting to this module, we construct $k-1$ ordered positive pairs $\left(v_{0}, v_{1}\right)$, ...\ , $\left(v_{k-2}, v_{k-1}\right)$, where $v_{i} = \mathcal{T}_{i}\left(s\right)$. 
Different from the previous works, the augmentations applied to one positive pair are different, 
\whzjh{which allows the model to be directional in its feature clustering.}
Meanwhile, we can also obtain the basic positive pair as described in Section~\ref{sec:skeletonclr}, $\left(v_{0}, v_{0}^{\prime}\right)$ via $\mathcal{T}_{0}$ (for $query$ and $key$).
The \whzjh{gradual growing} augmentation policy enables the model to treat augmentations differently by \whzjh{adjusting their applied branches} and \whzjh{decoupling} the learning of different augmentations.
\vspace{1mm}

\noindent \textbf{2) Asymmetric hierarchical learning.} Previous works based on contrastive learning utilize InfoNCE loss in Equation~\eqref{eq:infonce} for the representation learning.
However, it often leads to \whzjh{performance} drop when applying strong augmentations, which \whzjh{is caused by} the little mutual information \whzjh{among} different augmented views.
To this end, a hierarchical self-supervised learning objective is proposed to learn the representation consistency of multiple augmented views. 

As shown in Figure~\ref{fig:architecture}, the positive pairs are first encoded to the feature embeddings.
Formally, for a skeleton sequence $s$, we construct the positive pairs $\left(v_{0}, v_{1}\right)$, ...\ , $\left(v_{k-2}, v_{k-1}\right)$ and $\left(v_{0}, v_{0}^{\prime}\right)$ as discussed above. 
Then, the query encoder $f_{\theta_{q}}$ and the MLP head $h_{\theta_{q}}$ are applied successively to extract the feature representations: 
\begin{equation}
z_{i}=h_{\theta_{q}}\left(f_{\theta_{q}}\left(v_{i}\right)\right), i=0,1, ...\ ,k-1.
\end{equation}
Similarly, we can obtain the feature representation $z^{\prime}_{0}$ via the key encoder $f_{\theta_{k}}$ and the MLP $h_{\theta_{k}}$:
\begin{equation}
z^{\prime}_{0}=h_{\theta_{k}}\left(f_{\theta_{k}}\left(v^{\prime}_{0}\right)\right).
\end{equation}

The model optimizes the feature similarity of different augmented views $\left(v_{i-1}, v_{i}\right)$ to learn the representation consistency \whzjh{among} adjacent branches. 
\whzjh{Since these adjacent views share more augmentation strategies, it allows the target features to converge more smoothly to the center of the latent cluster.}
However, according to the previous work~\cite{bai2022directional}, it may lead to performance drop when the strongly augmented view is used as a mimic target due to the serious \whzjh{distortions}.
Therefore, we \whzjh{design} an asymmetric loss to unilaterally pull the features closer. 
The hierarchical self-supervised learning objective is computed by the feature similarity of adjacent branches and can be formulated as: 
\begin{equation}
\mathcal{L}_{h}=-\sum_{i=1}^{k-1}sim\left( z_{i}, {\rm stopgrad}\left(z_{i-1}\right)\right).
\end{equation}
Here, we utilize the stop-gradient ($\rm stopgrad$) operation to \whzjh{take} a more confident target for similarity learning. 
The strongly augmented view $z_{i}$ \whzjh{is} constrained to \whzjh{reduce} the feature distance from the weakly augmented view $z_{i-1}$, but not vice versa.
$sim(\cdot)$ can be any function that measures the similarity between two feature embeddings, such as cosine similarity and Kullback–Leibler (KL) divergence~\cite{kullback1951information}.
This can be viewed as an asymmetric design of representation consistency learning, which is adopted in Simsiam~\cite{chen2021exploring}, BYOL~\cite{grill2020bootstrap}, and CO2~\cite{wei2020co2}. 
Through the asymmetric hierarchical learning from multiple positive pairs, the model exploits the rich information brought by the strong augmentations and further improves the generalization capacity to downstream tasks.
\vspace{1mm}

\noindent\textbf{3) Instantiation.}
We next give an instantiation of our method.
For the asymmetric hierarchical learning, we use KL divergence as the $sim(\cdot)$ function. 
One problem is that it is difficult to calculate \whzjh{an} ideal accurate distribution of feature $z_{i}$. 
Inspired by ~\cite{wang2021contrastive}, we obtain the conditional distribution of feature $z_{i}$ with the positive feature output by the key encoder and numerous negative features maintained in \textbf{M}.
Specifically, the conditional distribution for $z_{i}$ is given as follows:
\begin{equation}\label{cond_distri}
	{p}\left({z|z_{i}}\right) = \frac{\exp(z \cdot z_{i} / \tau)}{\exp(z^{\prime}_{0} \cdot z_{i} / \tau) + \sum_{i=1}^{M}{\exp(m_{i} \cdot z_{i} / \tau)}}.
\end{equation}
\whzjh{Equation~\eqref{cond_distri}} depicts the similarity distribution of feature $z_{i}$ measured by positive features and negative features. 
According to Wang and Qi's discovery~\cite{wang2021contrastive}, the distributions of ${p}\left({z|z_{i}}\right)$ and ${p}\left({z|z_{i-1}}\right)$ \zjh{are similar via a randomly initialized network. It inspires us to optimize the distribution distances between ${p}\left({z|z_{i}}\right)$ and ${p}\left({z|z_{i-1}}\right)$,
{\ie,} $D_{KL}\left({\rm stopgrad}({p}\left({z|z_{i-1}}\right)), {p}\left({z|z_{i}}\right) \right)$ as $sim(\cdot)$, to learn the consistency between different augmented views.}
Also, we apply the $\mathcal{L}_{Info}$ on the basic positive pairs $\left(z_{0}, z^{\prime}_{0}\right)$ and jointly optimize the model. 
The overall loss is given by:
\begin{equation}
	\mathcal{L} = \mathcal{L}_{Info}+\lambda_{h}\mathcal{L}_{h},
\end{equation}
where $\lambda_{h}$ is the weight for hierarchical self-supervised loss.

For augmentations, the model is instantiated as $k$=3 with \textit{Basic Augmentation Set} (for $a_{0,*}$), \textit{Normal Augmentation Set} (for $a_{1,*}$) and \textit{Random Mask} (for $a_{2,*}$). 
We will discuss more about strong augmentations including \textit{Random Mask} in the next Section.

\subsection{Strong Augmentation for Skeleton}

We first introduce the \textit{Basic Augmentation Set} and \textit{Normal Augmentation Set} following the previous works~\cite{li20213d,guo2022aimclr,rao2021augmented}:
\begin{itemize}
    \item \noindent\textit{Basic Augmentation Set} (\textit{BA}) contains a spatial transformation \textit{Shear} and a temporal transformation \textit{Crop}.
    \item \noindent\textit{Normal Augmentation Set} (\textit{NA}) contains the following augments: \textit{Spatial Flip}, \textit{Rotation}, \textit{Gaussian Noise}, \textit{Gaussian Blur}, and \textit{Channel Mask}.
\end{itemize}

In addition to the augmentations above, we consider the strong augmentations targeted at skeleton data to introduce more novel patterns for the representation learning.
The augmentations for the skeleton are divided into three categories:
\begin{itemize}
    \item \textbf{Semantic-Dependent Augmentation.} Since human skeleton sequences have natural semantic information, we can perform linear transformations ({\eg,} rotation, scaling) or nonlinear transformations ({\eg,} joint replacement, flip) on 3D skeleton data to keep the essential semantic unchanged \zjh{and} computationally available.
    \item  \textbf{Feature-Wise Augmentation.} We can apply the disturbance to the features of graph nodes which is the human joints for skeleton data. It enables the model to obtain more robust representations for the noise \whzjh{in} collecting data, such as the error caused by the camera view.
    \item \textbf{Structure-Wise Augmentation.} Considering the topology of the human body, we hope that the model can output consistent semantic information under a slight perturbation of the joint adjacency graph. 
    It is because human action is often global and slight structural perturbations can be compensated by the information aggregation at other joints.
\end{itemize}

Based on this categorization, we propose the following three strong augmentation strategies as our \textit{Strong Augmentation Set} for skeleton:
\begin{itemize}
\item \textit{Random Mask.} A random mask for the spatial-temporal 3D coordinate data of the joints. It can be viewed as a random perturbation of the joint coordinates.

\item \textit{Drop/Add Edges (DAE)}. We
randomly drop/add connections between different joints in each information aggregation layer. The target to be augmented is the predefined or learnable adjacency matrix for the graph convolution layer and the attention map for the transformer block.

\item \textit{SkeleAdaIN.} Inspired by the practice of style transfer~\cite{huang2017arbitrary,karras2019style}, we exchange statistics of two skeleton samples on the spatial-temporal dimension, {\ie,} the mean and the variance of the style sample are transferred to the content sample, to generate the augmented views. Since this transformation does not change the relative order of joint coordinates, we maintain the semantics of skeleton sequences unchanged.
\end{itemize}

More details can be found in the Appendix.
These augmentations cause varying degrees of performance degradation when applied directly as shown on the right of Table~\ref{fig:h_loss}.
\whzjh{Therefore, we regard these augmentations as the examples of \textbf{strong augmentations} for the skeleton.}

\section{Experiment Results}
\subsection{Dataset}
\textbf{\whzjh{1)} NTU RGB+D Dataset 60 (NTU60)}~\cite{shahroudy2016ntu} is a large-scale dataset that contains 56,578 samples with 60 action categories and 25 joints. 
We follow the two recommended protocols: a) Cross-Subject (xsub): the data for training and testing are collected from different subjects. 
b) Cross-View (xview): the data for training and testing are collected from different camera views.

\noindent\textbf{\whzjh{2)} NTU RGB+D Dataset 120 (NTU120)}~\cite{liu2019ntu} is an extension to NTU60. 114,480 videos are collected with 120 action categories. Two recommended protocols are adopted: a) Cross-Subject (xsub): the data for training and testing are collected from 106 different subjects. b) Cross-Setup (xset)\whzjh{:} the data for training and testing are collected from 32 different setups.

\noindent\textbf{\whzjh{3)} PKU Multi-Modality Dataset (PKUMMD)}~\cite{liu2020benchmark} is a large-scale dataset covering \whzjh{a} multi-modality 3D understanding of human actions with almost 20,000 instances and 51 \whzjh{action} labels. Two subsets are divided: Part \uppercase\expandafter{\romannumeral1} is \whzjh{an} easier version; Part \uppercase\expandafter{\romannumeral2} provides more challenging data caused by large view variation and the cross-subject protocol is adopted.

\subsection{Implementation Details and Evaluation}
We evaluate the performance of our method on both ST-GCN~\cite{yan2018spatial} and DSTA-Net~\cite{shi2020decoupled} as backbones. The experimental settings of pre-training are following the previous works~\cite{li20213d,guo2022aimclr} for a fair comparison. All skeleton data are pre-processed into 50 frames. We reduce the number of channels in each graph convolution layer to 1/4 of the original setting for ST-GCN and 1/2 for DSTA-Net, respectively. The dimension of \whzjh{the} final output feature is 128 and the size of \whzjh{the} memory bank \textbf{M} is set to \whzjh{32,768}. The model is trained for 300 epochs with a batch-size of 128 using the SGD optimizer. $\lambda_{h}$ is set to 0.5. A multi-stream fusion strategy is adopted following the previous works, {\ie,} a weighted fusion of joint, bone, \whzjh{and} motion streams.
We adopt the following protocols to give a comprehensive evaluation:

\noindent\textbf{\whzjh{1)} KNN Evaluation.} 
We apply a K-Nearest Neighbor (KNN) classifier which is a non-parametric supervised learning method. It directly reflects the quality of the feature space learned by the encoder.

\noindent\textbf{\whzjh{2)} Linear Evaluation.}
A linear classifier is applied to the fixed encoder for linear evaluation. The classifier is trained to predict the corresponding label of the input sequences.

\noindent\textbf{\whzjh{3)} Semi-supervised Evaluation.}
In semi-supervised evaluation, we pre-train the encoder with all unlabeled data, and then train the whole model with randomly sampled 1\%, 10\% of the training data.

\noindent\textbf{\whzjh{4)} Supervised Evaluation.}
We fine-tune the whole model after pre-training the encoder. Both the encoder $f(\cdot)$ and classifier are trained for the downstream task.

\subsection{Ablation Study}
We first conduct ablation studies to give a more detailed analysis of our method. All results reported in this section are under linear evaluation on NTU60 dataset.
\vspace{1mm}

\begin{table}[t]
		\centering
		\small
		\renewcommand\arraystretch{0.9}{
		\begin{tabular}{l|c|cc}
			\toprule
			Augmentation  & Stream & xsub (\%)      & xview (\%)\\ 
			\midrule
			Baseline         &\multirow{4}{*}{Joint}       & 68.3 & 76.4\\
			Random Mask      &                             & \textbf{77.6} & 82.0\\
			DAE              &                             & 77.2 & 81.7\\ 
			SkeleAdaIN       &                             & 77.3 & \textbf{82.4}\\
			\midrule
			Baseline         &\multirow{4}{*}{Bone}        & 69.4 & 67.4\\
			Random Mask      &                             & 73.9 & 78.0\\
			DAE              &                             & \textbf{76.9} & \textbf{80.5}\\ 
			SkeleAdaIN       &                             & 75.3 & 79.2\\
			\midrule
			Baseline         &\multirow{4}{*}{Motion}      & 53.3 & 50.8\\
			Random Mask      &                             & 69.1 & \textbf{74.3}\\
			DAE              &                             & 68.0 & 72.2\\ 
			SkeleAdaIN       &                             & \textbf{69.5} & 71.8\\
			\midrule
			Baseline         &\multirow{4}{*}{Ensemble}    & 75.0 & 79.8\\
			Random Mask      &                             & \textbf{80.4} & \textbf{85.5}\\
			DAE              &                             & 79.8 & 84.9\\ 
			SkeleAdaIN       &                             & \textbf{80.4} & 84.4\\
			\bottomrule
		\end{tabular}
		}
		\vspace{-1mm}
		\caption{Ablation studies on the strong augmentations. Ensemble represents the fusion of joint-bone-motion streams.}
		\vspace{-5mm}
		\label{tab:data_aug}
\end{table}

\noindent \textbf{\whzjh{1)} Strong Augmentation Analysis.}
We set \textit{BA, NA} as the hierarchical augmentation set of the first and the second branch, and give an analysis when introducing the different strong augmentations as the extra augmentation for the third branch. The linear evaluation results are shown in Table~\ref{tab:data_aug}. Compared with the baseline, the model performance is significantly improved by applying the proposed strong augmentations with our HiCLR. Meanwhile, we also find some interesting results:

\whzjh{(a)} Different streams correspond to the different optimal augmentation methods. For example, \textit{DAE} performs significantly better than the other two augmentations on the bone stream. This may be relative to the association between invariances and streams, {\ie,} the bone view of the skeleton data implies the topological information of \whzjh{the} human body structure, which can be more robust to \textit{DAE} augmentation.

\whzjh{(b)} The performance of the same augmentation strategy can have a marked difference under different protocols. As shown in Table~\ref{tab:data_aug}, \textit{SkeleAdaIN} gives better results under cross-subject protocol than those under cross-view protocol. This is because \textit{SkeleAdaIN} can be regarded as a linear transformation of the action sequences under the same view, and the statistics which \whzjh{usually} \whzjh{contain} information about the performer's body shapes and range of motions are exchanged. Therefore, better robustness can be obtained under \whzjh{a} cross-subject evaluation protocol.

These results indicate that as a high-level representation, skeleton data faces more challenges in contrastive learning research. More efforts are needed in the design of augmentations for the skeleton data. To make our method more general, we finally adopt the \textit{Random Mask} augmentation in the third branch of our implementation. 

\begin{table}[t]
		\centering
		\small
		\renewcommand\arraystretch{0.9}{
        \begin{tabular}{l|c|cc}
			\toprule
			Arrangement & $k$ &xsub (\%) & xview (\%)\\ 
			\midrule
			${\text{[}}$BA, NA, Mask${\text{]}}$ &3  &77.6  &\textbf{82.0} \\
			${\text{[}}$NA, BA, Mask${\text{]}}$ &3  & \textbf{77.8} &80.3  \\ 
			${\text{[}}$Mask, BA, NA${\text{]}}$ &3 & {74.7} & 79.7\\
			${\text{[}}$BA+NA, Mask${\text{]}}$  &2 &74.0  &79.0 \\
			${\text{[}}$BA, NA+Mask${\text{]}}$  &2 &76.7  &79.4 \\
			\bottomrule
		\end{tabular}
		}
		\vspace{-1mm}
		\caption{Ablation studies on the data augmentation arrangement of the single joint stream.}
		\label{tab:arrange}
\end{table}

\begin{table}[t]
\begin{minipage}[t]{0.36\linewidth}
    \vspace{0pt}
    \small
    \renewcommand{\arraystretch}{0.9}
    \begin{tabular}{l|c}
			\toprule
			\textit{$sim(\cdot)$}  &Acc. \\ 
			\midrule
			Cosine & 75.8\%  \\
			L1                            & 73.6\%   \\ 
			KL div.                 & \textbf{77.6\%} \\
			\bottomrule
		\end{tabular}
\hfill
\end{minipage}
  \begin{minipage}[t]{0.58\linewidth}
    \vspace{0pt}
    \small
    \renewcommand{\arraystretch}{0.9}
    \setlength{\tabcolsep}{1.2mm}{
    \begin{tabular}[c]{l|c|c}
 \toprule
  Augmentation & Baseline & Ours \\
  \midrule
  ${\text{[}}$BA${\text{]}}$ & 68.3\% & -  \\
 \midrule
 ${\text{[}}$BA, NA${\text{]}}$ & 72.9\%  & \textbf{76.8\%} \\
\midrule
${\text{[}}$BA, NA, Mask${\text{]}}$ & 56.7\% & \textbf{77.6\%} \\
${\text{[}}$BA, NA, DAE${\text{]}}$ & 65.5\% & \textbf{77.2\%} \\
${\text{[}}$BA, NA, AdaIN${\text{]}}$ & {13.2\%} & \textbf{77.3\%} \\
\bottomrule
\end{tabular}}
\end{minipage}
\vspace{-1mm}
\caption{The accuracy is reported under cross-subject protocol. Left: The effect of different similarity functions in hierarchical self-supervised loss. Right: Ablation studies of the hierarchical design when applying different augmentations.}
\vspace{-5mm}
\label{fig:h_loss}
\end{table}

\noindent \textbf{\whzjh{2)} Data Augmentation Arrangement.}
Table~\ref{tab:arrange} shows the results of different augmentation arrangements, where BA, NA, and Mask represent the \textit{Basic}, \textit{Normal Augmentation Set}, and \textit{Random Mask} augmentation, respectively. As we can see, different arrangements can have a marked influence on the results which demonstrates the necessity of making a discriminate treatment for different augmentations. It is found that the optimal method approximates a kind of arrangement from weak augmentations to strong augmentations. This also \whzjh{proves} that strong augmentation is not suitable as the basic augmentation strategy, confirming our hierarchical learning ideas from easy to difficult.

\noindent \textbf{\whzjh{3)} Hierarchical Consistent Learning.}
\label{doc:loss_abl}
As shown on the left of Table~\ref{fig:h_loss}, KL divergence gives the best results as $sim(\cdot)$ function, indicating that the distribution of ${p}\left({z|z_{i}}\right)$ and ${p}\left({z|z_{i-1}}\right)$ should be similar for a well-pre-trained model~\cite{wang2021contrastive}. It can be regarded as a soft version of InfoNCE loss, which introduces more samples to measure and constrain the consistency of different augmented views. Meanwhile, the results when more and strong augmentations are applied are shown on the right of Table~\ref{fig:h_loss}. As we can see, HiCLR can bring a consistent improvement even though some augmentations such as \textit{Random Mask}, \textit{DAE}, and \textit{SkeleAdaIN} show adverse effects on the baseline algorithm, verifying the effectiveness of HiCLR.

\begin{table*}[t]
	\centering
	\small
	\renewcommand\arraystretch{0.9}{
	\begin{tabular}{l|c|c|cc|cc}
	
		\toprule
		\multirow{2}{*}{Method}&\multirow{2}{*}{Backbone} &\multirow{2}{*}{Params} &\multicolumn{2}{c|}{NTU60}&\multicolumn{2}{c}{NTU120}\\
			  &  & &xsub (\%)& xview (\%)&xsub (\%)&xset (\%)\\ 
		\midrule
		LongT GAN (AAAI 18) &GRU  &40.2M  &39.1 &48.1  & - & -\\
		MS$\rm^{2}$L (ACM MM 20) &GRU &2.28M  &52.6   &-    & - & -\\
		P\&C (CVPR 20) &GRU  &-   &50.7  &76.3   & 42.7 & 41.7\\ 
		AS-CAL (Information Sciences 21) & LSTM &0.43M  &58.5 &64.8 & 48.6 &49.2 \\
		ISC (ACM MM 21) &{GRU+GCN+CNN} &10.0M &76.3 &85.2 &67.1 &67.9\\
		SeBiReNet (ECCV 20) &GRU &0.27M &- &79.7 & - & -\\
		\midrule
		3s-CrosSCLR (CVPR 21) &ST-GCN &0.85M  &77.8 &83.4 & 67.9 & 66.7\\
		3s-AimCLR (AAAI 22)  &ST-GCN  & 0.85M &78.9 &83.8 &68.2  & 68.8 \\
		\textbf{Ours} &ST-GCN &\textbf{0.85M} &\textbf{80.4} &\textbf{85.5}  &\textbf{70.0}  &\textbf{70.4}\\
		\midrule
		H-Transformer (ICME 21) &Transformer &\textgreater 100M &69.3 &72.8 &- & -\\
		GL-Transformer (ECCV 22) &Transformer &214M &76.3 & \textbf{83.8} &66.0 & 68.7\\
		\textbf{Ours} &Transformer &\textbf{1.56M} &\textbf{78.8} &83.1 &\textbf{67.3} &\textbf{69.9} \\
		\bottomrule
	
	\end{tabular}
	}
	\vspace{-1mm}
	\caption{Linear evaluation results on NTU60 and NTU120 datasets.}
	\vspace{-4mm}
	\label{tab:ntu_linear}
\end{table*}
	
\begin{table}[t]
		\centering
		\small
		\renewcommand\arraystretch{0.9}{
		\begin{tabular}{l|c|cc|c}
			\toprule
			\multirow{2}{*}{Method} &\multirow{2}{*}{Stream}&\multicolumn{2}{c|}{NTU60 (\%)}&{PKUMMD}\\
			& & xsub& xview &Part \uppercase\expandafter{\romannumeral1} (\%)\\ 
			\midrule
			SkeletonCLR&\multirow{3}{*}{Joint}   & 56.1 & 61.7&68.9\\
			AimCLR     &                         & 62.0 &71.5&72.0 \\ 
			\textbf{Ours}        &                & \textbf{67.3} & \textbf{75.3}&\textbf{73.8}\\
			\midrule
			SkeletonCLR&\multirow{3}{*}{Motion}   & 37.4 & 41.6 &51.0\\
			AimCLR   &      & 50.8 & 56.9 &60.6\\ 
			\textbf{Ours} &                & \textbf{55.3} & \textbf{60.7} & \textbf{63.8}\\
			\bottomrule
		\end{tabular}
		}
		\vspace{-1mm}
		\caption{KNN evaluation results of different streams.}
		\vspace{-2mm}
		\label{tab:knn_res}
\end{table}

\begin{table}[t]
		\centering
		\small
		\renewcommand\arraystretch{0.9}{
		\begin{tabular}{l|cc|cc}
			\toprule
			\multirow{2}{*}{Method} &\multicolumn{2}{c|}{\textit{1\% data}}&\multicolumn{2}{c}{\textit{10\% data}}\\
			 & xsub & xview& xsub& xview\\
			\midrule
			ASSL (20)&- &- & 64.3 &69.8 \\ 
			MCC (21) &- &- & 55.6 & 59.9\\
			3s-CrosSCLR (21) &51.1 &50.0 &74.4 &77.8\\
			3s-Colorization (21)&48.3 &52.5 &71.7 &78.9\\
			3s-AimCLR (22)&54.8 &54.3 &78.2 &81.6\\
			\textbf{Ours} (GCN)&\textbf{58.5} &\textbf{58.3} &\textbf{79.6} &\textbf{84.0}\\
			\midrule
			3s-Hi-TRS (22)&49.3 &51.5 &77.7 &81.1\\
			\textbf{Ours} (Transformer) &\textbf{54.7}   &\textbf{53.7}   &\textbf{82.1}   &\textbf{84.8}\\
			\bottomrule
		\end{tabular}
		}
		\vspace{-1mm}
		\caption{Semi-supervised results on \whzjh{NTU60} dataset.}
		\vspace{-4mm}
		\label{tab:semi_res}
\end{table}
\begin{table}[t]
		\centering
		\small
		\renewcommand\arraystretch{0.9}{
		\begin{tabular}{l|c|cc}
			\toprule
			{Method} &{Params}&\multicolumn{2}{c}{Protocol}\\
			\midrule
			 \textcolor{gray}{NTU60 Dataset} &  &\textcolor{gray}{xsub}&\textcolor{gray}{xview}\\
			
			3s-CrosSCLR (CVPR 21)&0.85M &86.2 &92.5\\
			3s-AimCLR (AAAI 22)&0.85M &86.9 &92.8\\
			\textbf{Ours} (GCN)&0.85M &\textbf{88.3} &\textbf{93.2}\\
			\midrule
			3s-Hi-TRS (ECCV 22)&7.05M &90.0 &\textbf{95.7}\\
			\textbf{Ours} (Transformer)&\textbf{1.56M} &\textbf{90.4} &\textbf{95.7}\\
			\midrule
			\textcolor{gray}{NTU120 Dataset} &  &\textcolor{gray}{xsub}&\textcolor{gray}{xset}\\
			3s-CrosSCLR (CVPR 21)&0.85M &80.5 &80.4\\
			3s-AimCLR (AAAI 22)&0.85M &80.1 &80.9\\
			\textbf{Ours} (GCN)&0.85M &\textbf{82.1} &\textbf{83.7}\\
			\midrule
			3s-Hi-TRS (ECCV 22)&7.05M &85.3 &87.4\\
			\textbf{Ours} (Transformer)&\textbf{1.56M} &\textbf{85.6} &\textbf{87.5} \\
			\bottomrule
		\end{tabular}
		}
		\vspace{-1mm}
		\caption{Supervised results on NTU \whzjh{dataset}.}
		\vspace{-4mm}
		\label{tab:finetune_ntu60}
\end{table}

\subsection{Comparison with State-of-the-art Methods}
We compare our method with the state-of-the-art methods for self-supervised skeleton-based action recognition under different evaluation protocols.
\vspace{1mm}

\label{doc:data_aug}
\noindent \textbf{\whzjh{1)} Linear Evaluation Results.} We use both GCNs and transformers as our backbone to comprehensively demonstrate the effectiveness of our approach.
First, compared with other GCN-based methods~\cite{zheng2018unsupervised,lin2020ms2l,su2020predict,rao2021augmented,thoker2021skeleton,nie2020unsupervised,li20213d,guo2022aimclr,yang2021skeleton}, HiCLR has achieved the best performance on NTU datasets as shown in Table~\ref{tab:ntu_linear}. By virtue of the hierarchical design, our method benefits better from strong augmentations and significantly outperforms the results of other methods. Compared with AimCLR~\cite{guo2022aimclr}, which also considers the strong augmented views, we obtain a notable improvement on both the single joint stream (\textbf{77.6\%} vs. 74.3\% on xsub and \textbf{82.0\%} vs. 79.7\% on xview) and the fusion results.

We also compare the latest works using transformers~\cite{cheng2021hierarchical,Kim2022GloballocalMT} as shown in Table~\ref{tab:ntu_linear}. 
HiCLR uses only one percent of the model parameters to achieve comparable or better performance than others, indicating the efficiency and effectiveness of our method. 

\noindent \textbf{\whzjh{2)} KNN Evaluation Results.} The KNN evaluation is a direct reflection of the quality of the feature space~\cite{wu2018unsupervised}. In Table~\ref{tab:knn_res}, we can see that our method outperforms the SkeletonCLR and AimCLR by a large margin on both the joint and motion streams. It indicates that a higher quality feature space is learned by the model owing to the introduction of more strong augmentations. 

\noindent \textbf{\whzjh{3)} Semi-supervised Evaluation Results.}
The semi-supervised results are presented in Table~\ref{tab:semi_res}. Our method can significantly improve the performance in semi-supervised learning compared with GCN-based methods~\cite{si2020adversarial,su2021self}, especially when there is little training data available. 
Meanwhile, a remarkable gain is observed when using transformers as the backbone. Compared with Hi-TRS~\cite{chen2022hierarchically}, HiCLR improves the semi-supervised results by a large margin, verifying the strong representation ability of our method.

\noindent \textbf{\whzjh{4)} Supervised Evaluation Results.}
We conduct supervised evaluation experiments and the results are shown in Table~\ref{tab:finetune_ntu60}. Compared with other GCN-based methods, HiCLR consistently outperforms other methods, especially on NTU120 dataset. Moreover, for the latest transformer-based method, our method can exceed Hi-TRS~\cite{chen2022hierarchically} with fewer parameters and renews the state-of-the-art score.

\section{Conclusion}
In this paper, we propose a new hierarchical contrastive learning framework, HiCLR, to fully take advantage of the strong augmentations. 
Instead of learning all augmentations \whzjh{without distinction}, HiCLR \whzjh{learns from} hierarchical consistency with growing augmentations, alleviating the difficulty \whzjh{in} learning consistency from the strongly augmented views. 
An asymmetric loss is applied to \whzjh{align the feature extracted from the strongly augmented view to the one from the weakly augmented view}. 
Extensive experiments \whzjh{verify} the effectiveness of HiCLR \whzjh{for} GCNs and \whzjh{transformers} \whzjh{as backbones}.
HiCLR can generate a more distinguishable feature space \whzjh{and} outperforms the state-of-the-art methods under \whzjh{various} protocols.
\label{sec:reference_examples}


\vspace{.2em}

\bibliography{aaai23}
\end{document}